\documentclass{article} 
\usepackage{iclr2020_conference,times}

\usepackage{amsmath,amsfonts,bm}

\def\1{\bm{1}}

\DeclareMathAlphabet{\mathsfit}{\encodingdefault}{\sfdefault}{m}{sl}
\SetMathAlphabet{\mathsfit}{bold}{\encodingdefault}{\sfdefault}{bx}{n}

\usepackage{multicol}
\usepackage{color}
\usepackage{amssymb}
\usepackage{amsmath}
\usepackage{amsthm}
\usepackage{tikz}
\usepackage{bbm}
\usetikzlibrary{arrows,decorations.markings}
\usepackage[makeroom]{cancel}

\usepackage{booktabs}
\usepackage{graphicx}
\usepackage{todonotes}

\usepackage[font=small]{caption}
\usepackage{subcaption}

\usepackage[noline]{algorithm2e}
\makeatletter
\newcommand{\removelatexerror}{\let\@latex@error\@gobble}
\makeatother

\usepackage{indentfirst}
\usepackage{float}
\usepackage{mathtools}

\newenvironment{tightlist}%
{\begin{list}{$\bullet$}{%
    \setlength{\topsep}{0in}
    \setlength{\partopsep}{0in}
    \setlength{\itemsep}{0in}
    \setlength{\parsep}{0in}
    \setlength{\leftmargin}{1.5em}
    \setlength{\rightmargin}{0in}
    \setlength{\itemindent}{-.1in}
}
}%
{\end{list}
}

\newcommand{\secref}[1]{Section~\ref{#1}}

\newcommand{\eqnref}[1]{Equation~\ref{#1}}
\newcommand{\figref}[1]{Figure~\ref{#1}}

\newcommand{\appenref}[1]{Appendix~\ref{#1}}

\DeclarePairedDelimiterX{\infdivx}[2]{(}{)}{%
  #1\;\delimsize\|\;#2%
}

\newcommand{\St}{\mathcal{S}}
\newcommand{\A}{\mathcal{A}}

\newcommand{\ppo}{{\sc ppo}}
\newcommand{\mdp}{{\sc mdp}}

\newcommand{\M}{\mathcal{M}}

\newcommand{\Env}{\mathcal{E}}
\newcommand{\data}{\mathcal{D}}

\newcommand{\Ex}{\mathbb{E}}

\usepackage{hyperref}
\usepackage{lib}

\title{Intrinsic Motivation for Encouraging\\ Synergistic Behavior}

\author{
  \textbf{Rohan Chitnis\thanks{Work done during an internship at Facebook AI Research.} \hspace{12mm} Shubham Tulsiani \hspace{12mm} Saurabh Gupta \hspace{12mm} Abhinav Gupta}\\\\
  \small{MIT Computer Science and Artificial Intelligence Laboratory, Facebook Artificial Intelligence Research}\\
  \texttt{\small{ronuchit@mit.edu, shubhtuls@fb.com, saurabhg@illinois.edu, gabhinav@fb.com}}
}

\newcommand{\fjoint}{{f^\text{joint}}}
\newcommand{\rone}{{r_1^{\text{intrinsic}}}}
\newcommand{\rtwo}{{r_2^{\text{intrinsic}}}}

\iclrfinalcopy 
\begin{document}

\maketitle

\begin{abstract}
  We study the role of intrinsic motivation as an exploration bias for
  reinforcement learning in sparse-reward synergistic tasks,
  which are tasks where multiple agents must work together to achieve a goal
  they could not individually. Our key idea is that a good guiding
  principle for intrinsic motivation in synergistic tasks is to take
  actions which affect the world in ways that would not be achieved
  if the agents were acting on their own. Thus, we propose to incentivize
  agents to take (joint) actions whose effects cannot be predicted via
  a composition of the predicted effect for each individual agent. We
  study two instantiations of this idea, one based on the true states
  encountered, and another based on a dynamics model trained
  concurrently with the policy. While the former is simpler, the
  latter has the benefit of being analytically differentiable with
  respect to the action taken. We validate our approach in robotic
  bimanual manipulation and multi-agent locomotion tasks with sparse rewards; we find that our
  approach yields more efficient learning than both 1) training with
  only the sparse reward and 2) using the typical surprise-based
  formulation of intrinsic motivation, which does not bias toward
  synergistic behavior.  Videos are available on the project webpage:
  \href{https://sites.google.com/view/iclr2020-synergistic}{\texttt{https://sites.google.com/view/iclr2020-synergistic}}.
\end{abstract}

\section{Introduction}
\label{sec:intro}
Consider a multi-agent environment such as a team of robots working
together to play soccer. It is critical for a joint policy within such
an environment to produce synergistic behavior, allowing multiple
agents to work together to achieve a goal which they could not achieve
individually. How should agents learn such synergistic behavior
efficiently? A naive strategy would be to learn policies jointly and
hope that synergistic behavior emerges. However, learning policies
from \emph{sparse, binary rewards} is very challenging -- exploration
is a huge bottleneck when positive reinforcement is infrequent and
rare. In sparse-reward multi-agent environments where synergistic
behavior is critical, exploration is an even bigger issue due to the
much larger action space.

A common approach for handling the exploration bottleneck in reinforcement learning is to shape
the reward using \emph{intrinsic motivation}, as was first proposed by
\citet{schmidhuberintrinsic}. This has been shown to yield improved
performance across a variety of domains, such as robotic control
tasks~\citep{oudeyerintrinsic} and Atari games~\citep{unifying,
  deepakcuriosity}. Typically, intrinsic motivation is formulated as
the agent's prediction error regarding some aspects of the world;
shaping the reward with such an error term incentivizes the agent to
take actions that ``surprise it,'' and is intuitively a
useful heuristic for exploration. But is this a good strategy for
encouraging synergistic behavior in multi-agent settings? Although
synergistic behavior may be difficult to predict, it could be equally
difficult to predict the effects of certain single-agent behaviors;
this formulation of intrinsic motivation as ``surprise'' does not
specifically favor the emergence of synergy.

In this paper, we study an alternative strategy for employing
intrinsic motivation to encourage synergistic behavior in multi-agent
tasks. Our method is based on the simple insight that synergistic
behavior leads to effects which would not be achieved if the
individual agents were acting alone. So, we propose to reward agents
for joint actions that lead to different results compared to if those
same actions were done by the agents individually, in a sequential
composition. For instance, consider the task of twisting open a water
bottle, which requires two hands (agents): one to hold the base in
place, and another to twist the cap. Only holding the base in place
would not effect any change in the bottle's pose, while twisting the
cap without holding the bottle in place would cause the entire bottle
to twist, rather than just the cap. Here, holding with one hand and
\emph{subsequently} twisting with the other would not open the bottle,
but holding and twisting \emph{concurrently} would.

Based on this intuition, we propose a formulation for intrinsic
motivation that leverages the difference between the true effect of an
action and the composition of individual-agent predicted effects. We
then present a second formulation that instead uses the discrepancy of
predictions between a joint and a compositional prediction
model. While the latter formulation requires training a forward model
alongside learning the control strategy, it has the benefit of being
analytically differentiable with respect to the action taken. We later
show that this can be leveraged within the policy gradient framework,
in order to obtain improved sample complexity over using the policy
gradient as-is.

\begin{figure}[t]
\begin{center}
\includegraphics[width=0.9\columnwidth]{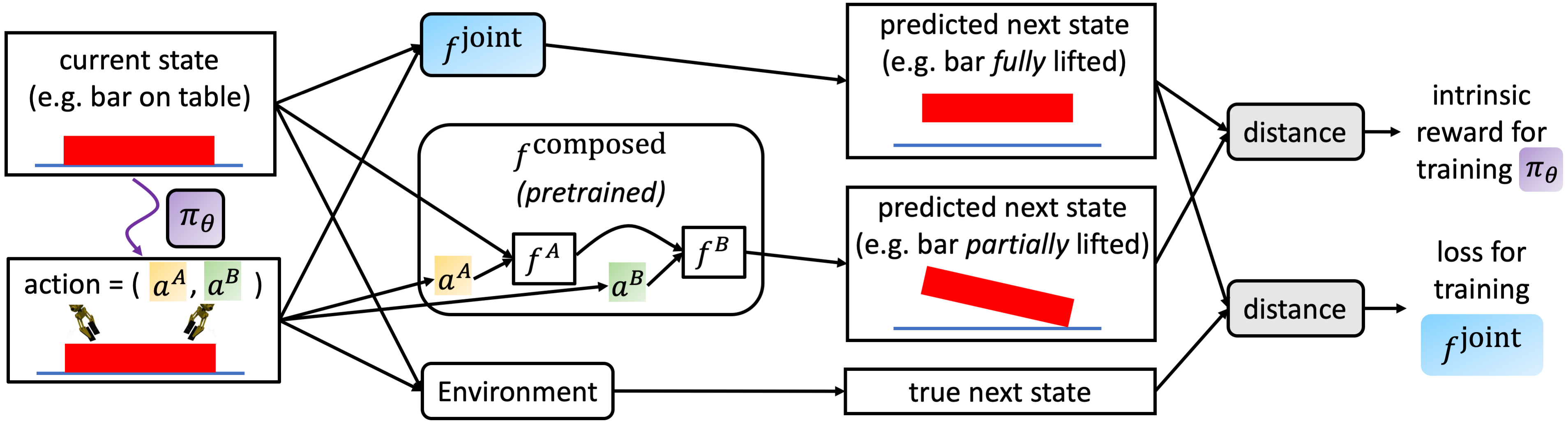}
\end{center}
\caption{An overview of our approach to incentivizing synergistic
  behavior via intrinsic motivation. A heavy red bar (requiring two
  arms to lift) rests on a table, and the policy $\pi_\theta$ suggests
  for arms $A$ and $B$ to lift the bar from opposite ends. A
  composition of pretrained single-agent forward models, $f^A$ and
  $f^B$, predicts the resulting state to be one where the bar is only
  partially lifted, since neither $f^A$ nor $f^B$ has ever encountered
  states where the bar is lifted during training. A forward model
  trained on the complete two-agent environment, $f^{\text{joint}}$,
  correctly predicts that the bar is fully lifted, very different from
  the compositional prediction. We train $\pi_\theta$ to prefer
  actions such as these, as a way to bias toward synergistic
  behavior. Note that differentiating this intrinsic reward with
  respect to the action taken does \emph{not} require differentiating
  through the environment.}
\label{fig:overview}
\vspace{-1em}
\end{figure}

As our experimental point of focus, we study six simulated robotic
tasks: four bimanual manipulation (bottle opening, ball pickup,
corkscrew rotating, and bar pickup) and two multi-agent locomotion
(ant push and soccer). All tasks have sparse rewards: 1 if the goal
is achieved and 0 otherwise. These tasks were chosen both because
they require synergistic behavior, and because they represent
challenging control problems for modern state-of-the-art deep
reinforcement learning
algorithms~\citep{visuomotor,ddpg,naf,a3c,mbmf}. Across all tasks, we
find that shaping the reward via our formulation of intrinsic
motivation yields more efficient learning than both 1) training with
only the sparse reward signal and 2) shaping the reward via the more
standard single-agent formulation of intrinsic motivation as
``surprise,'' which does not explicitly encourage synergistic
behavior. We view this work as a step toward general-purpose
synergistic multi-agent reinforcement learning.

\section{Related Work}
\label{sec:rw}
\textbf{Prediction error as intrinsic motivation.} The idea of
motivating an agent to reach areas of the state space which yield high
model prediction error was first proposed by
\citet{schmidhuberintrinsic}. Generally, this reward obeys the form
$\|f(x)-\hat{f}(x)\|$, i.e. the difference between the predicted and
actual value of some function computed on the current state, the taken
action, etc.~\citep{barto2013intrinsic,oudeyerintrinsic,unifying};
intrinsic motivation can even be used on its own when no extrinsic
reward is
provided~\citep{deepakcuriosity,disagreement,largescalecuriosity,haber2018emergence}. A
separate line of work studies how agents can synthesize a library of
\emph{skills} via intrinsic motivation in the absence of extrinsic
rewards~\citep{diayn}. Recent work has also studied the use of
surprise-based reward to solve gentle manipulation tasks, with the
novel idea of rewarding the agent for errors in its own predictions of
the reward function~\citep{sandycuriosity}. In this paper, we will
propose formulations of intrinsic motivation that are geared toward
multi-agent synergistic tasks.

\textbf{Exploration in multi-agent reinforcement learning.} The
problem of efficient exploration in multi-agent settings has received
significant attention over the years. Lookahead-based
exploration~\citep{lookaheadexplore} is a classic strategy; it rewards
an agent for exploration that reduces its uncertainty about the models
of other agents in the environment. More recently, \emph{social
motivation} has been proposed as a general principle for guiding
exploration~\citep{socialmotivation}: agents should prefer actions
that most strongly influence the policies of other
agents. \textsc{lola}~\citep{lola}, though not quite an exploration
strategy, follows a similar paradigm: an agent should reason about the
impact of its actions on how other agents learn.  Our work approaches
the problem from a different angle that incentivizes synergy: we
reward agents for taking actions to affect the world in ways that
would not be achieved if the agents were acting alone.

\textbf{Bimanual manipulation.} The field of bimanual, or dual-arm,
robotic manipulation has a rich history~\citep{bimanualsurvey} as an
interesting problem across several areas, including hardware design,
model-based control, and reinforcement learning. Model-based control
strategies for this task often draw on hybrid force-position control
theory~\citep{hybridcontrol}, and rely on analytical models of the
environment dynamics, usually along with assumptions on how the
dynamics can be approximately decomposed into terms corresponding to
the two arms~\citep{hybridcontroldual1,hybridcontroldual2}. On the
other hand, learning-based strategies for this task often leverage
human demonstrations to circumvent the challenge of
exploration~\citep{lfddual1,lfddual2,lfddual3}. In this work, we
describe an exploration strategy based on intrinsic motivation.

\section{Approach}
\label{sec:approach}
Our goal is to enable learning for synergistic tasks in settings with
sparse extrinsic rewards. A central hurdle in such scenarios is the
exploration bottleneck: there is a large space of possible action
sequences that the agents must explore in order to see rewards. In
the absence of intermediate extrinsic rewards to guide this
exploration, one can instead rely on \emph{intrinsic} rewards that
bias the exploratory behavior toward ``interesting'' actions, a notion
which we will formalize.

To accomplish any synergistic task, the agents must work together to
affect the environment in ways that would not occur if they were
working individually. In \secref{sec:form1}, we present a formulation
for intrinsic motivation that operationalizes this insight and allows
guiding the exploration toward synergistic behavior, consequently
learning the desired tasks more efficiently. In \secref{sec:form2}, we present a second formulation that is (partially)
differentiable, making learning even more
efficient by allowing us to compute analytical gradients with respect
to the action taken. Finally, in \secref{sec:rl} we show how our
formulations can be used to efficiently learn task policies.

\textbf{Problem Setup.}  Each of the tasks we consider can be
formulated as a two-agent finite-horizon \mdp~\citep{mdp}.\footnote{Our problem setup and proposed approach can be extended to
settings with more than two agents. Details, with accompanying
experimental results, are provided in \secref{subsec:3agents}.}  We denote
the environment as $\Env$, and the agents as $A$ and $B$. We assume a
state $s \in \St$ can be partitioned as
$s \vcentcolon= \langle s^A, s^B, s^{\text{env}} \rangle$, where
$s^A \in \St^A$, $s^B \in \St^B$, and
$s^{\text{env}} \in \St^{\text{env}}$. Here, $s^A$ and $s^B$ denote
the proprioceptive states of the agents, such as joint configurations
of robot arms, and $s^{\text{env}}$ captures the remaining aspects of
the environment, such as object poses. An action $a \in \A$ is a tuple
$a \vcentcolon= \langle a^A, a^B \rangle$, where $a^A \in \A^A$ and
$a^B \in \A^B$, consisting of each agent's actions.

We focus on settings where the reward function of this \mdp\ is binary
and sparse, yielding reward $r^{\text{extrinsic}}(s) = 1$ only when
$s$ achieves some desired goal configuration. Learning in such a setup
corresponds to acquiring a (parameterized) policy $\pi_\theta$ that
maximizes the expected proportion of times that a goal configuration
is achieved by following $\pi_\theta$.

Unfortunately, exploration guided only by a sparse reward is
challenging; we propose to additionally bias it via an intrinsic
reward function. Let $\bar{s} \sim \Env(s, a)$ be a next state
resulting from executing action $a$ in state $s$. We wish to formulate
an intrinsic reward function $r^{\text{intrinsic}}(s, a, \bar{s})$
that encourages synergistic actions and can thereby enable more
efficient learning.

\subsection{Compositional Prediction Error as an Intrinsic Reward}
\label{sec:form1}
We want to encourage actions that affect the environment in ways that
would not occur if the agents were acting individually. To formalize
this notion, we note that a ``synergistic'' action is one where the
agents acting \emph{together} is crucial to the outcome; so, we should
expect a different outcome if the corresponding actions were executed
\emph{sequentially}, with each individual agent acting at a time.

Our key insight is that we can leverage this difference between the
\emph{true outcome} of an action and the \emph{expected outcome with
  individual agents acting sequentially} as a reward signal. We can
capture the latter via a composition of forward prediction models for
the effects of actions by individual agents acting
separately. Concretely, let
$f^A: \St^{\text{env}} \times \St^A \times \A^A \to \St^{\text{env}}$
(resp. $f^B$) be a single-agent prediction model that regresses to the
next environment state resulting from $A$ (resp. $B$) taking an action
in isolation.\footnote{As the true environment dynamics are
  stochastic, it can be useful to consider probabilistic regressors
  $f$. However, recent successful applications of model-based
  reinforcement learning~\citep{mbmf,clavera2018learning} have used
  deterministic regressors, modeling just the maximum likelihood
  transitions.}
We define our first formulation of intrinsic reward, $r_1^{\text{intrinsic}}(s, a, \bar{s})$, by measuring the prediction error of $\bar{s}^{\text{env}}$ using a composition of these single-agent prediction models:
\begin{gather*}
    f^{\text{composed}}(s, a) =  f^B( f^A(s^{\text{env}}, s^A, a^A), s^B, a^B),\\
    r_1^{\text{intrinsic}}(s, a, \bar{s}) = \|\bar{s}^{\text{env}} - f^{\text{composed}}(s,a)\|.
\end{gather*}
For synergistic actions $a$, the prediction $f^{\text{composed}}(s, a)$ will likely be quite different from $\bar{s}^{\text{env}}$.

In practice, we pretrain $f^A$ and $f^B$ using data of random
interactions in instantiations of the environment $\Env$ with only a
single active agent. This implies that the agents have already
developed an understanding of the effects of acting alone before being
placed in multi-agent environments that require synergistic
behavior. Note that while random interactions sufficed to learn useful
prediction models $f^A$ and $f^B$ in our experiments, this is not
essential to the formulation, and one could leverage alternative
single-agent exploration strategies to collect interaction samples
instead.

\subsection{Prediction Disparity as a Differentiable Intrinsic Reward}
\label{sec:form2}
The reward $r_1^{\text{intrinsic}}(s, a, \bar{s})$ presented above
encourages actions that have a synergistic effect. However, note that
this ``measurement of synergy'' for action $a$ in state $s$ requires
explicitly observing the outcome $\bar{s}$ of executing $a$ in the
environment. In contrast, when humans reason about synergistic tasks
such as twisting open a bottle cap while holding the bottle base, we
judge whether actions will have a synergistic effect without needing
to execute them to make this judgement. Not only is the non-dependence
of the intrinsic reward on $\bar{s}$ scientifically interesting, but
it is also practically desirable. Specifically, the term
$f^{\text{composed}}(s,a)$ is analytically differentiable with respect
to $a$ (assuming that one uses differentiable regressors $f^A$ and
$f^B$, such as neural networks), but $\bar{s}^{\text{env}}$ is not,
since $\bar{s}$ depends on $a$ via the black-box environment. If we
can reformulate the intrinsic reward to be analytically differentiable
with respect to $a$, we can leverage this for more sample-efficient
learning.

To this end, we observe that our formulation rewards actions where the
expected outcome under the compositional prediction differs from the
outcome when the agents act together. While we used the observed state
$\bar{s}$ as the indication of ``outcome when the agents act
together,'' we could instead use a \emph{predicted} outcome here. We
therefore additionally train a \emph{joint} prediction model
$f^{\text{joint}}: \St \times \A \to \St^{\text{env}}$ that, given the
states and actions of \emph{both} agents, and the environment state,
predicts the next environment state. We then define our second
formulation of intrinsic reward,
$r_2^{\text{intrinsic}}(s, a, \cdot)$, using the disparity between the
predictions of the joint and compositional models:
\begin{gather*}
    r_2^{\text{intrinsic}}(s, a, \cdot) = \|f^{\text{joint}}(s, a) - f^{\text{composed}}(s, a)\|.
\end{gather*}

Note that there is no dependence on $\bar{s}$. At first, this
formulation may seem less efficient than $r_1^{\text{intrinsic}}$,
since $f^{\text{joint}}$ can at best only match
$\bar{s}^{\text{env}}$, and requires being trained on data. However,
we note that this formulation makes the intrinsic reward analytically
differentiable with respect to the action $a$ executed; we can leverage this within the
learning algorithm to obtain more informative gradient updates, as we
discuss further in the next section.

\noindent {\bf Relation to Curiosity.} Typical approaches to intrinsic
motivation~\citep{stadie2015incentivizing,deepakcuriosity}, which
reward an agent for ``doing what surprises it,'' take on the form
$r_\text{non-synergistic}^{\text{intrinsic}}(s, a, \bar{s}) =
\|f^{\text{joint}}(s, a) - \bar{s}^{\text{env}}\|$. These
curiosity-based methods will encourage the system to keep finding new
behavior that surprises it, and thus can be seen as a technique for
curiosity-driven skill discovery.  In contrast, we are focused on
synergistic multi-agent tasks with an extrinsic (albeit sparse)
reward, so our methods for intrinsic motivation are not intended to
encourage a diversity of learned behaviors, but rather to bias
exploration to enable sample-efficient learning for a given task.

\subsection{Learning Sparse-Reward Synergistic Tasks}
\label{sec:rl}
We simultaneously learn the joint prediction model $f^{\text{joint}}$
and the task policy $\pi_{\theta}$. We train $\pi_\theta$ via
reinforcement learning to maximize the expected total shaped reward
$r^{\text{full}} = r^{\text{intrinsic}}_{i \in \{1, 2\}}+\lambda\cdot
r^{\text{extrinsic}}$ across an episode. Concurrently, we make
dual-purpose use of the transition samples $\{(s, a, \bar{s})\}$
collected during the interactions with the environment to train
$f^{\text{joint}}$, by minimizing the loss
$\|f^{\text{joint}}(s, a) - \bar{s}^{\text{env}}\|$. This simultaneous
training of $f^{\text{joint}}$ and $\pi_{\theta}$, as was also done by
\citet{stadie2015incentivizing}, obviates the need for collecting
additional samples to pretrain $f^{\text{joint}}$ and ensures that the
joint prediction model is trained using the ``interesting''
synergistic actions being explored. Full pseudocode is provided in
\appenref{app:pseu}.

Our second intrinsic reward formulation allows us to leverage
differentiability with respect to the action taken to make learning
via policy gradient methods more efficient. Recall that any policy
gradient algorithm~\citep{ppo,trpo,reinforce} performs gradient ascent
with respect to policy parameters $\theta$ on the expected reward over
trajectories:
$J(\theta) \vcentcolon= \Ex_\tau[r^{\text{full}}(\tau)]$. Expanding,
we have
$J(\theta) = \Ex_\tau\left[\sum_{t=0}^T r^{\text{full}}(s_t, a_t,
  \cdot)\right] = \Ex_\tau\left[\sum_{t=0}^T
  r_2^{\text{intrinsic}}(s_t, a_t, \cdot)+\lambda\cdot
  r^{\text{extrinsic}}(s_t) \right]$, where $T$ is the horizon. We
show in \appenref{app:derive} that the gradient can be written as:
\begin{gather}
 \nabla_\theta J(\theta) =\sum_{t=0}^T \Ex_{\tau_t}[r^{\text{full}}(s_t,a_t,\cdot)\nabla_\theta \log p_\theta(\bar{\tau}_t)]+\Ex_{\bar{\tau}_t}[\nabla_\theta \Ex_{a_t \sim \pi_{\theta}(s_t)} [r^{\text{intrinsic}}_2(s_t,a_t,\cdot)]].
 \label{eq:gradient}
\end{gather}
Here, $\tau_t \vcentcolon= \langle s_0, a_0, ..., s_t, a_t \rangle$
denotes a trajectory up to time $t$, and
$\bar{\tau}_t \vcentcolon= \langle s_0, a_0, ..., s_t \rangle$
denotes the same but \textit{excluding} $a_t$. Given a state $s_t$,
and assuming a differentiable way of sampling
$a_t \sim \pi_{\theta}(s_t)$, such as using the reparameterization
trick~\citep{kingma2013auto}, we can analytically compute the inner
gradient in the second term since
$r_2^{\text{intrinsic}}(s_t, a_t, \cdot)$ is differentiable with
respect to $a_t$ (again, assuming the regressors $f^A$, $f^B$, and
$f^{\text{joint}}$ are differentiable). In \eqnref{eq:gradient}, the first
term is similar to what typical policy gradient algorithms compute,
with the difference being the use of $p_\theta(\bar{\tau}_t)$ instead
of $p_\theta(\tau_t)$; the intuition is that we should not consider
the effects of $a_t$ here since it gets accounted
for by the second term. In practice, however, we opt to treat the
policy gradient algorithm as a black box, and simply add (estimates
of) the gradients given by the second term to the gradients yielded by
the black-box algorithm. While this leads to double-counting certain gradients (those of the expected reward at each timestep with respect to the action at that timestep), our preliminary
experiments found this to minimally affect training, and make the
implementation more convenient as one can leverage an off-the-shelf
optimizer like \ppo~\citep{ppo}.

\section{Experiments}
\label{sec:experiments}
We consider both bimanual
manipulation tasks and multi-agent locomotion tasks, all of which
require synergistic behavior, as our testbed. We establish the utility
of our proposed formulations by comparing to baselines that do not use
any intrinsic rewards, or use alternative intrinsic reward
formulations. We also consider ablations of our method that help us
understand the different intrinsic reward formulations, and the impact
of partial differentiability. In \secref{subsec:3agents}, we show that our
approach, with minor adaptations, continues to be useful in domains with
more than two agents.

\subsection{Experimental Setup}
We consider four bimanual manipulation tasks: bottle opening, ball
pickup, corkscrew rotating, and bar pickup. These environments
are suggested as bimanual manipulation tasks by~\citet{chitnisschemas}. Furthermore,
we consider two multi-agent locomotion tasks: ant push (inspired by the domain considered by~\citet{diversification}) and
soccer (adapted from the implementation provided
alongside~\citet{emergent}). All tasks involve sparse rewards, and
require effective use of both agents to be solved. We simulate all
tasks in MuJoCo~\citep{mujoco}. Now, we describe the tasks, state
representations, and action spaces.

\begin{figure}[t]
\begin{center}
\includegraphics[width=0.24\linewidth]{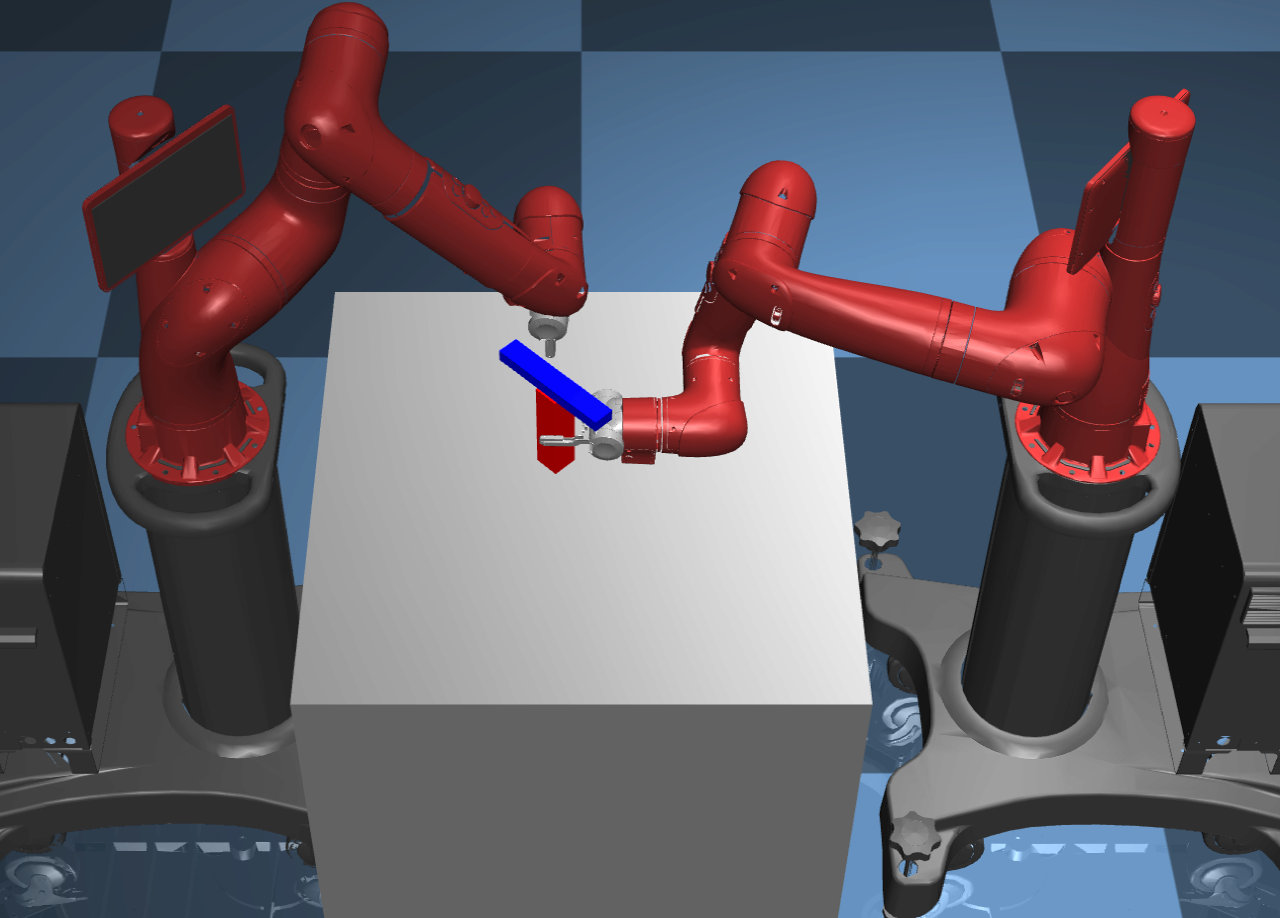}
\includegraphics[width=0.24\linewidth]{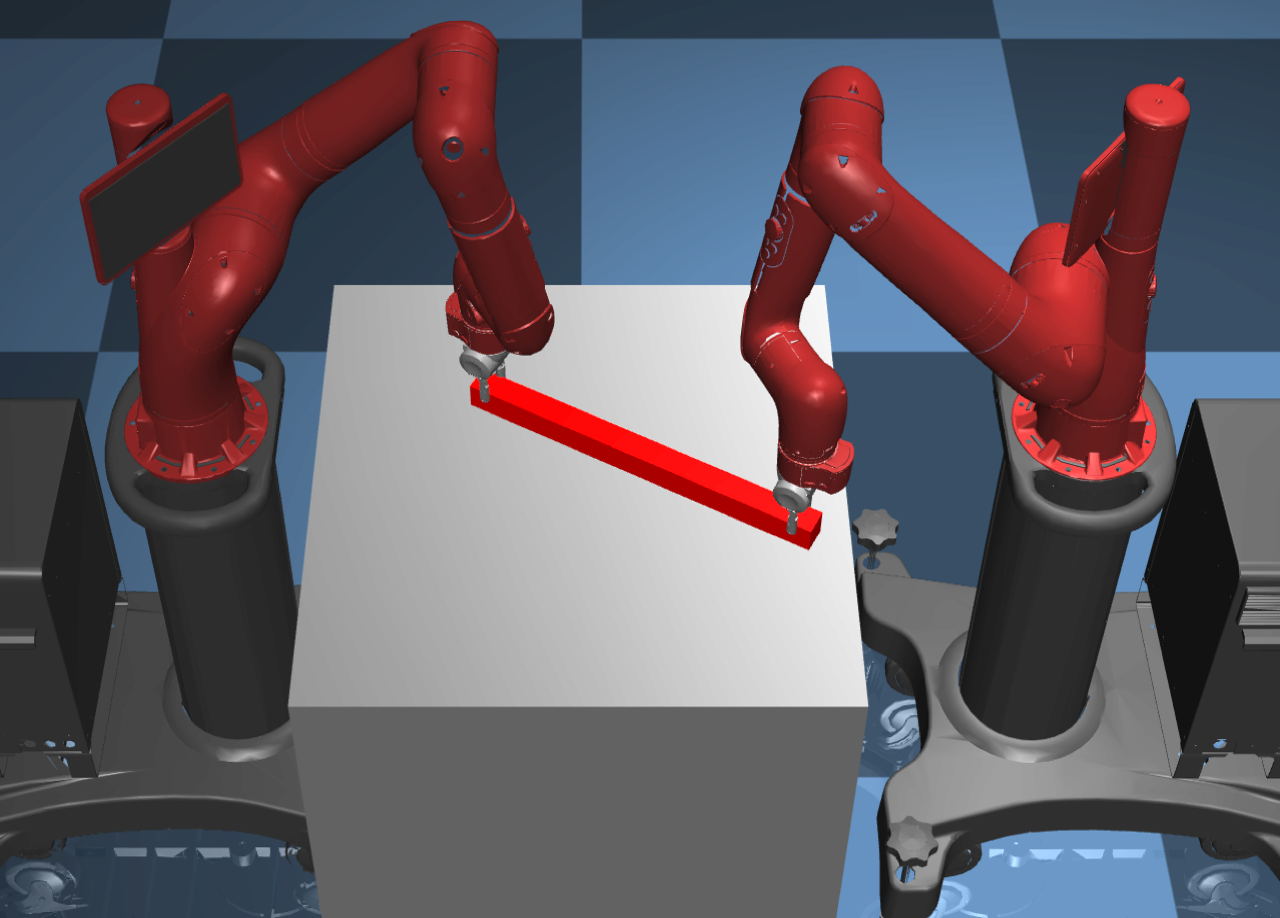}
\includegraphics[width=0.24\linewidth]{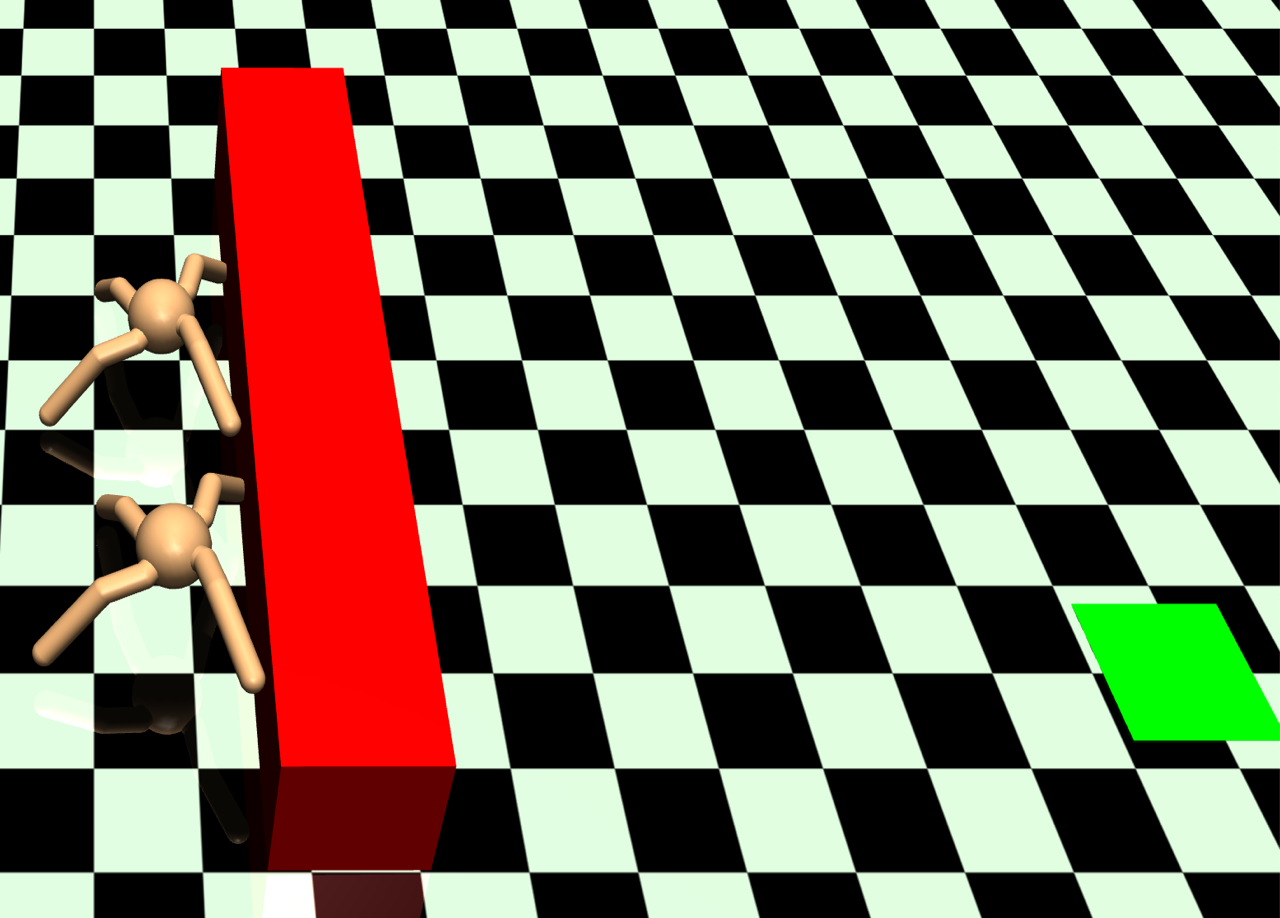}
\includegraphics[width=0.24\linewidth]{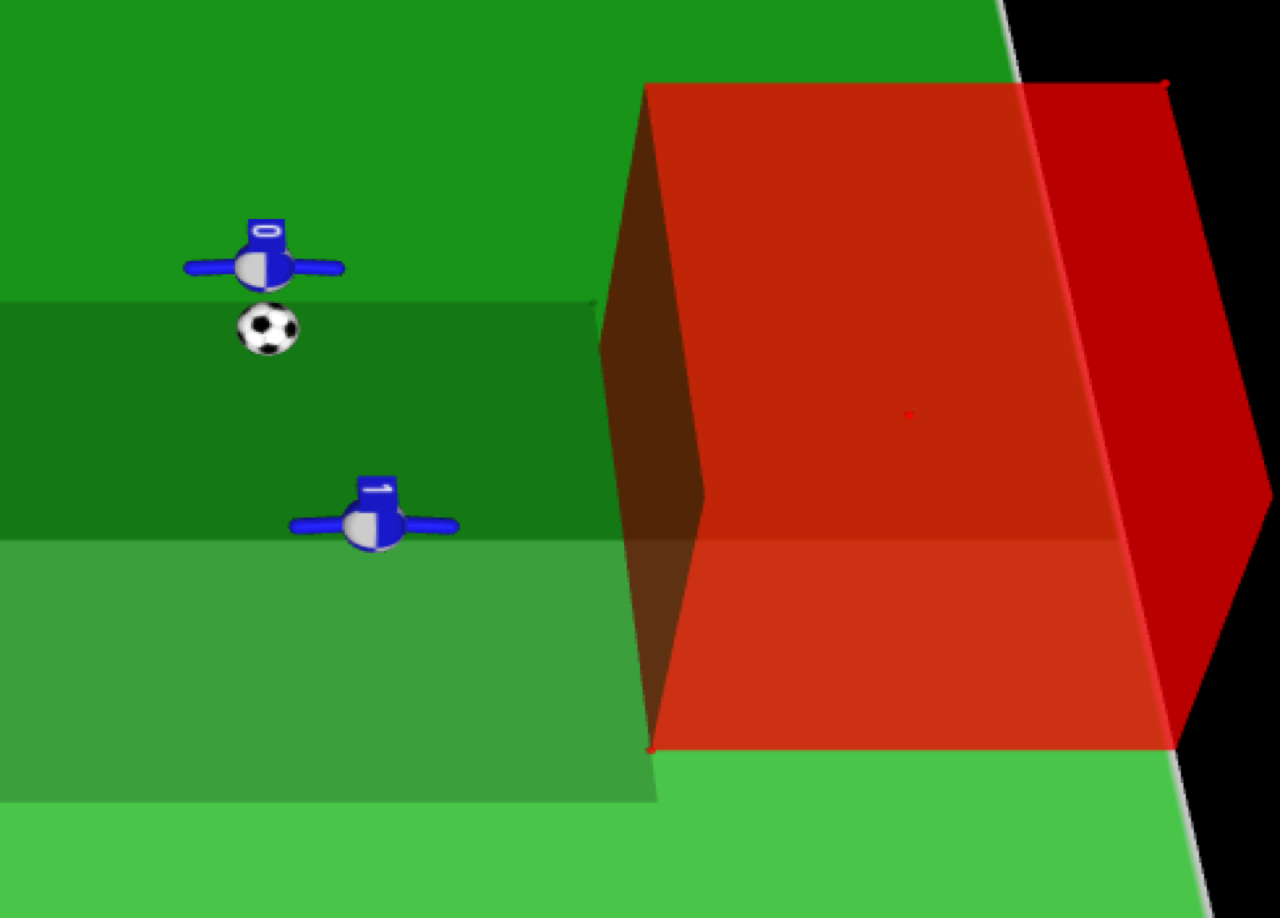}
\end{center}
\caption{Screenshots of two of our manipulation tasks and our
  locomotion tasks. From left to right: corkscrew rotating, bar
  pickup, ant push, soccer. These tasks are all designed to require
  two agents. We learn policies for these tasks given only sparse
  binary rewards, by encouraging synergistic behavior via intrinsic
  motivation.}
\label{fig:screenshots}
\end{figure}

\noindent \textbf{Environments.}  The four manipulation tasks are set
up with 2 Sawyer arms at opposite ends of a table, and an object
placed on the table surface. Two of these tasks are visualized in
\figref{fig:screenshots}, alongside the two multi-agent locomotion
tasks.
\begin{tightlist}
\item \textit{Bottle Opening}: The goal is to rotate a cuboidal bottle
  cap, relative to a cuboidal bottle base, by $90^\circ$. The bottle
  is modeled as two cuboids on top of one another, connected via a
  hinge joint, such that in the absence of opposing torques, both
  cuboids rotate together. We vary the location and size of the bottle
  across episodes.
\item \textit{Ball Pickup}: The goal is to lift a slippery ball by
  25cm. The ball slips out when a single arm tries to lift it. We vary
  the location and coefficient of friction of the ball across
  episodes.
\item \textit{Corkscrew Rotating}: The goal is to rotate a corkscrew
  relative to its base by $180^\circ$. The corkscrew is modeled as a
  handle attached to a base via a hinge joint, such that in the
  absence of opposing torques, both rotate together. We vary the
  location and size of the corkscrew across episodes.
\item \textit{Bar Pickup}: The goal is to lift a long heavy
  bar by 25cm. The bar is too heavy to be lifted by a single arm. We
  vary the location and density of the bar across episodes.
\item \textit{Ant Push}: Two ants and a large block are placed in an
  environment. The goal is for the ants to move the block to a particular
  region. To control the block precisely, the ants need to push it
  together, as they will often topple over when trying to push the
  block by themselves.
\item \textit{Soccer}: Two soccer-playing agents and a soccer ball are
  placed in an environment. The goal is for the ball to be kicked into
  a particular region, \emph{after} having been in the possession of each
  agent for any amount of time. Therefore, the agents must both
  contribute to the movement of the ball.
\end{tightlist}
See \secref{subsec:3agents} for results on three-agent versions of the
Ant Push and Soccer environments.

\noindent \textbf{State Representation.} The internal state of each
agent consists of proprioceptive features: joint positions, joint
velocities, and (for manipulation tasks) the end effector pose. The
environment state consists of the current timestep, geometry
information for the object, and the object pose. We use a simple
Euclidean metric over the state space. All forward models predict the
change in the object's world frame pose, via an additive offset for
the 3D position and a Hamilton product for the orientation
quaternion. The orientation is not tracked in the soccer task.

\noindent \textbf{Action Space.} To facilitate learning within these
environments, we provide the system with a discrete library of generic
skills, each parameterized by some (learned) continuous
parameters. Therefore, our stochastic policy $\pi_\theta$ maps a state
to 1) a distribution over skills for agent $A$ to use, 2) a
distribution over skills for agent $B$ to use, 3) means and variances
of independent Gaussian distributions for every continuous parameter
of skills for $A$, and 4) means and variances of independent Gaussian
distributions for every continuous parameter of skills for $B$. These
skills can either be hand-designed~\citep{primitives1,primitives2} or
learned from demonstration~\citep{lfddual3}; as this is not the focus
of our paper, we opt to simply hand-design them. While executing a
skill, if the agents are about to collide with each other, we attempt
to bring them back to the states they were in before
execution. For manipulation tasks, if we
cannot find an inverse kinematics solution for achieving a skill, it
is not executed, though it still consumes a timestep. In either of
these cases, the reward is 0. See \appenref{app:envdetails} for more
details on these environments.

\subsection{Implementation Details}
\noindent \textbf{Network Architecture.} All forward models and the policy are 4-layer fully connected neural networks with 64-unit hidden layers, ReLU activations, and a multi-headed output to capture both the actor and the critic. Bimanual manipulation tasks are built on the Surreal Robotics Suite~\citep{robosuite}. For all tasks, training is parallelized across 50 workers.

\noindent \textbf{Training Details.} Our proposed synergistic
intrinsic rewards rely on forward models $f^A$, $f^B$, and
$\fjoint$. We pretrain the single-agent model $f^A$ (resp. $f^B$) on
$10^5$ samples of experience with a random policy of only agent $A$
(resp. $B$) acting. Note that this pretraining does not use any
extrinsic reward, and therefore the number of steps under the
extrinsic reward is comparable across all the approaches. The joint
model $f^{\text{joint}}$ and policy $\pi_\theta$ start from scratch,
and are optimized concurrently. We set the trade-off coefficient
$\lambda=10$ (see \appenref{app:lambda}). We use the stable baselines~\citep{stable-baselines}
implementation of \ppo~\citep{ppo}
as our policy gradient algorithm. We use clipping parameter 0.2,
entropy loss coefficient 0.01, value loss function coefficient 0.5,
gradient clip threshold 0.5, number of steps 10, number of minibatches
per update 4, number of optimization epochs per update 4, and
Adam~\citep{adam} with learning rate 0.001.

\subsection{Baselines}

\begin{tightlist}
\item \textit{Random policy}: We randomly choose a skill and
  parameterization for each agent, at every step. This baseline serves
  as a sanity check to ensure that our use of skills does not
  trivialize the tasks.
\item \textit{Separate-agent surprise}: This baseline simultaneously
  executes two independent single-agent curiosity policies that are
  pretrained to maximize the ``surprise'' rewards
  $\|f^A(s,a)-\bar{s}^{\text{env}}\|$ and
  $\|f^B(s,a)-\bar{s}^{\text{env}}\|$ respectively.
\item \textit{Extrinsic reward only}: This baseline uses only
  extrinsic sparse rewards $r^{\text{extrinsic}}$, without shaping.
\item \textit{Non-synergistic surprise}: We learn a joint two-agent
  policy to optimize for the extrinsic reward and the joint surprise:
  $r^{\text{full}}=r_{\text{non-synergistic}}^{\text{intrinsic}}+\lambda
  \cdot r^{\text{extrinsic}}$. This encourages curiosity-driven skill
  discovery but does not explicitly encourage synergistic multi-agent
  behavior.
\end{tightlist}

\begin{figure}[t]
\begin{center}
\includegraphics[width=\linewidth]{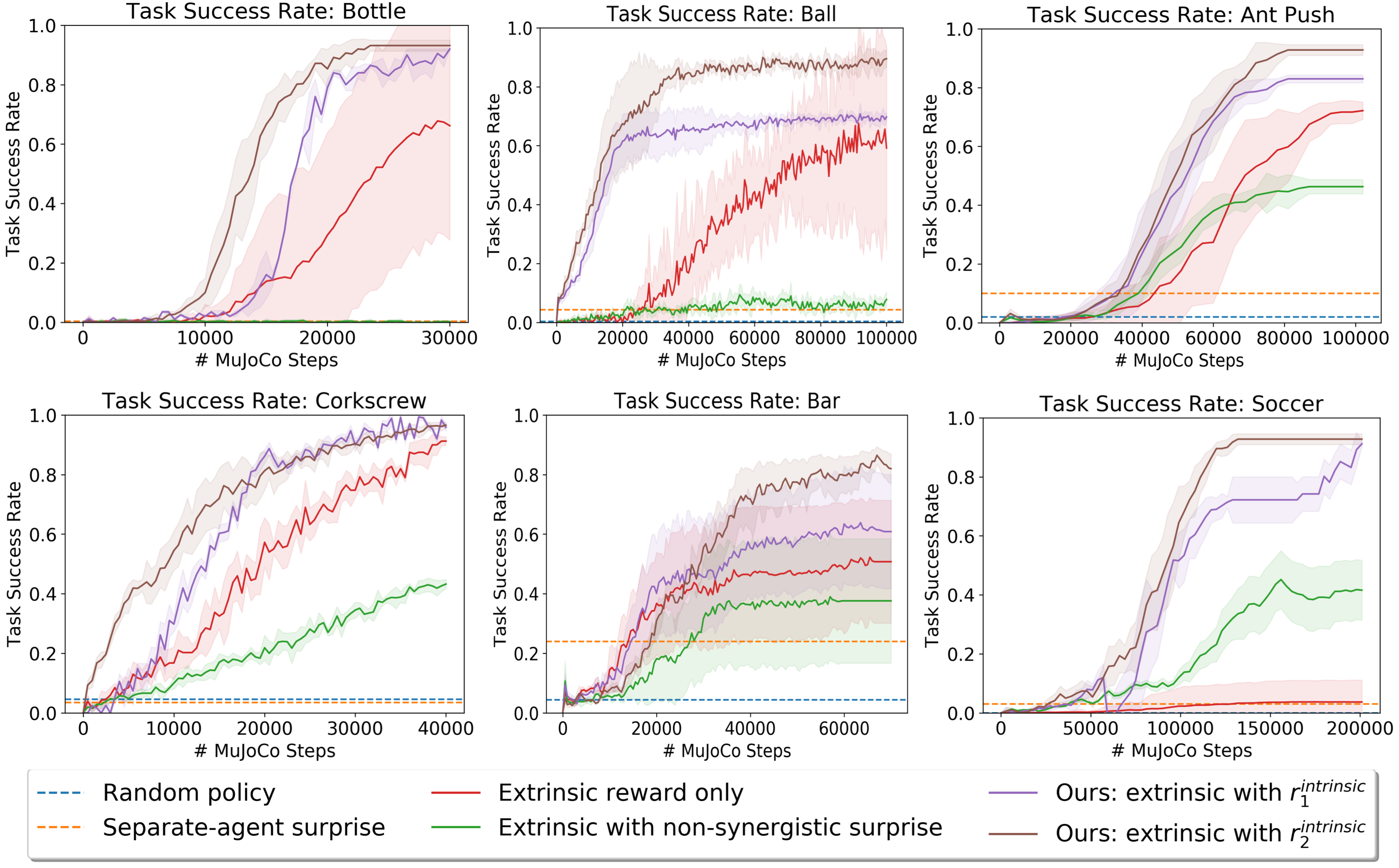}
\end{center}
\caption{Learning curves for each of our environments. Each curve
  depicts an average across 5 random seeds, with standard deviations shaded. We see that it is
  much more sample-efficient to shape the reward via
  $r_1^{\text{intrinsic}}$ or $r_2^{\text{intrinsic}}$ than to rely
  only on the extrinsic, sparse reward signal. Also, typical
  formulations of intrinsic motivation as surprise do not work well
  for synergistic tasks because they encourage the
  system to affect the environment in ways it cannot currently
  predict, while our approach encourages the system to affect the
  environment in ways neither agent would if acting on its own, which is
  a useful bias for learning synergistic behavior.}
\label{fig:curves}
\end{figure}

\subsection{Results and Discussion}
\figref{fig:curves} shows task success rates as a function of the
number of interaction samples for the different methods on each
environment. We plot average success rate over 5 random seeds using
solid lines, and shade standard deviations. Now, we summarize our
three key takeaways.

\textbf{1) Synergistic intrinsic rewards boost sample efficiency.} The
tasks we consider are hard and our use of parameterized skills does
not trivialize the tasks. Furthermore, these tasks require
coordination among the two agents, and so \textit{Separate-agent surprise}
policies do not perform well. Given enough training samples,
\textit{Extrinsic reward only} policies start to perform decently
well. However, our use of synergistic intrinsic rewards to shape the
extrinsic rewards from the environment accelerates learning, solving
the task consistently with up to $5\times$ fewer samples in some
cases.

\begin{figure}[t]
\begin{center}
\includegraphics[width=\linewidth]{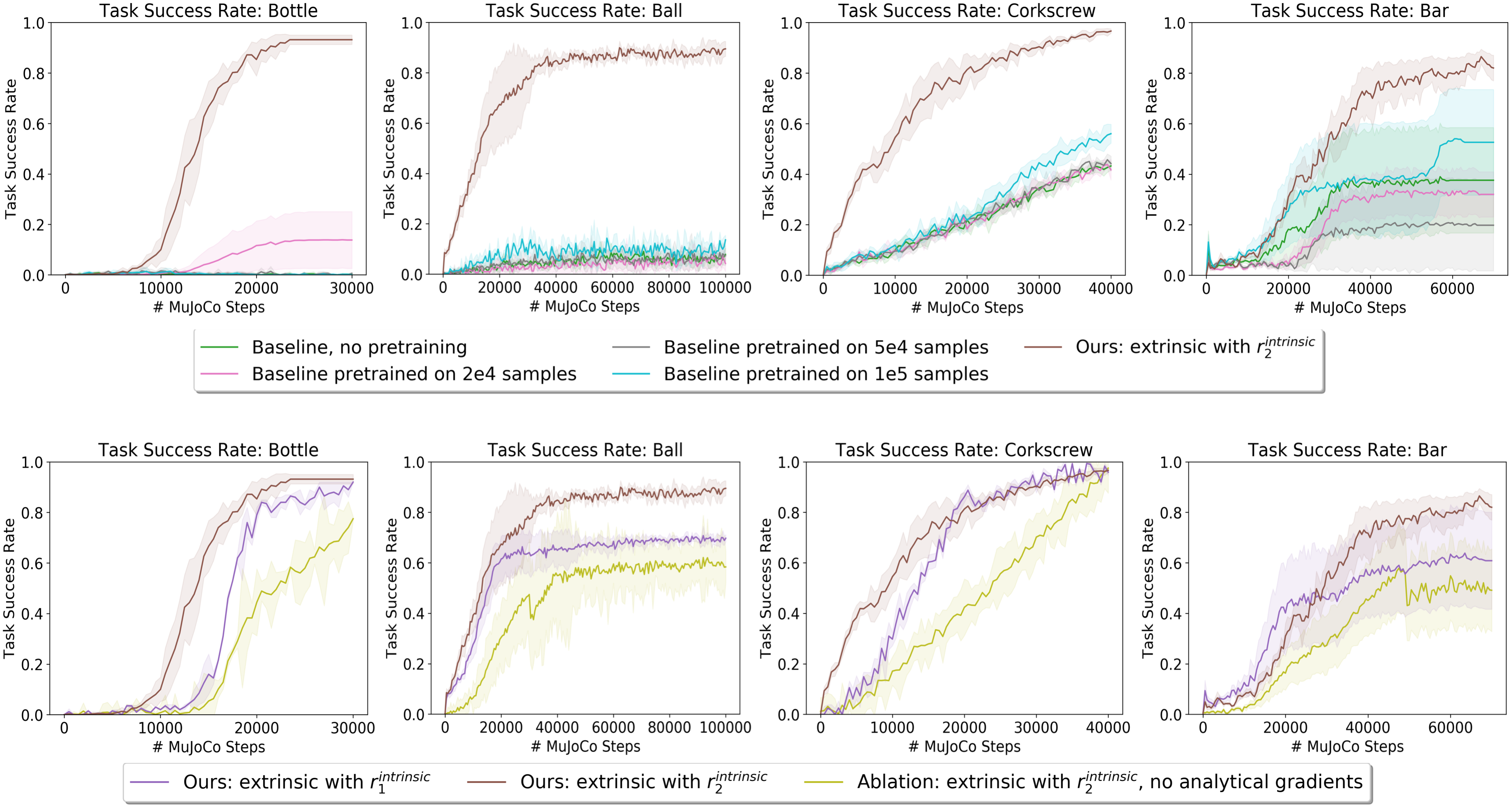}
\end{center}
\caption{\emph{Top}: \textit{Non-synergistic surprise} baseline with
  varying amounts of pretraining for the joint model
  $f^{\text{joint}}$. We see that pretraining this joint model does
  not yield much improvement in performance, and remains significantly
  worse than our method (brown curve). This is sensible since the
  baseline does not explicitly encourage synergistic behavior, as we
  do. \emph{Bottom}: Ablation showing the impact of using analytical
  gradients on sample efficiency. $\rtwo$ only performs better than
  $\rone$ when leveraging the partial differentiability.}
\label{fig:othercurves}
\end{figure}

\textbf{2) Synergistic intrinsic rewards perform better than
  non-synergistic intrinsic rewards.} Policies that use our
synergistic intrinsic rewards also work better than the
\textit{Non-synergistic surprise} baseline. This is primarily because
the baseline policies learn to exploit the joint model rather than to
behave synergistically. This also explains why \textit{Non-synergistic
  surprise} used together with extrinsic reward hurts task performance
(green vs. red curve in \figref{fig:curves}). Past experiments with
such surprise models have largely been limited to games, where
progress is correlated with continued
exploration~\citep{largescalecuriosity}; solving robotic tasks often
involves more than just surprise-driven
exploration. \figref{fig:othercurves} (top) gives additional results
showing that our method's competitive advantage over this baseline
persists even if we allow the baseline additional interactions to
pretrain the joint prediction model $f^{\text{joint}}$ without using
any extrinsic reward (similar to our method's pretraining for
$f^{\text{composed}}$).

\textbf{3) Analytical gradients boost sample efficiency.} In going
from $\rone$ (compositional prediction error) to $\rtwo$ (prediction
disparity), we changed two things: 1) the reward function and 2) how
it is optimized (we used \eqnref{eq:gradient} to leverage the partial
differentiability of $\rtwo$). We conduct an ablation to disentangle
the impact of these two changes. \figref{fig:othercurves} (bottom)
presents learning curves for using $\rtwo$ \textit{without} analytical
gradients, situated in comparison to the previously shown
results. When we factor out the difference due to optimization and
compare $\rone$ and $\rtwo$ as different intrinsic reward
formulations, $\rone$ performs better than $\rtwo$ (purple vs. yellow
curve). This is expected because $\rtwo$ requires training an extra
model $f^{\text{joint}}$ concurrently with the policy, which at best
could match the true $\bar{s}^{\text{env}}$. Leveraging the analytical
gradients, though, affords $\rtwo$ more sample-efficient optimization
(brown vs. purple curve), making it a better overall choice.

We have also tried using our formulation of intrinsic motivation
\emph{without} extrinsic reward ($\lambda=0$); qualitatively, the
agents learn to act synergistically, but in ways that do not solve the
``task,'' which is sensible since the task is unknown to the
agents. See the project webpage for videos of these
results. Furthermore, in \appenref{app:lambda} we provide a plot of
policy performance versus various settings of $\lambda$.

\subsection{Extension: More than Two Agents}
\label{subsec:3agents}
It is possible to extend our formulation and proposed approach
to more than two agents. Without loss of generality, suppose there are
three agents $A$, $B$, and $C$. The only major change is in the way
that we should compute the compositional prediction: instead of
$f^{\text{composed}}(s, a) = f^B( f^A(s^{\text{env}}, s^A, a^A), s^B,
a^B)$, we use
$f^{\text{composed}}(s, a) = f^C ( f^B( f^A(s^{\text{env}}, s^A, a^A),
s^B, a^B), s^C, a^C)$.
One issue is that as the number of agents increases, the ordering of
the application of single-agent forward models within
$f^{\text{composed}}$ becomes increasingly important. To address this,
we also tried evaluating $f^{\text{composed}}$ as an average across
the predictions given by all six possible orderings of application,
but we did not find this to make much difference in the results. We
leave a thorough treatment of this important question to future work.

\begin{figure}[h]
\begin{center}
\includegraphics[width=0.8\linewidth]{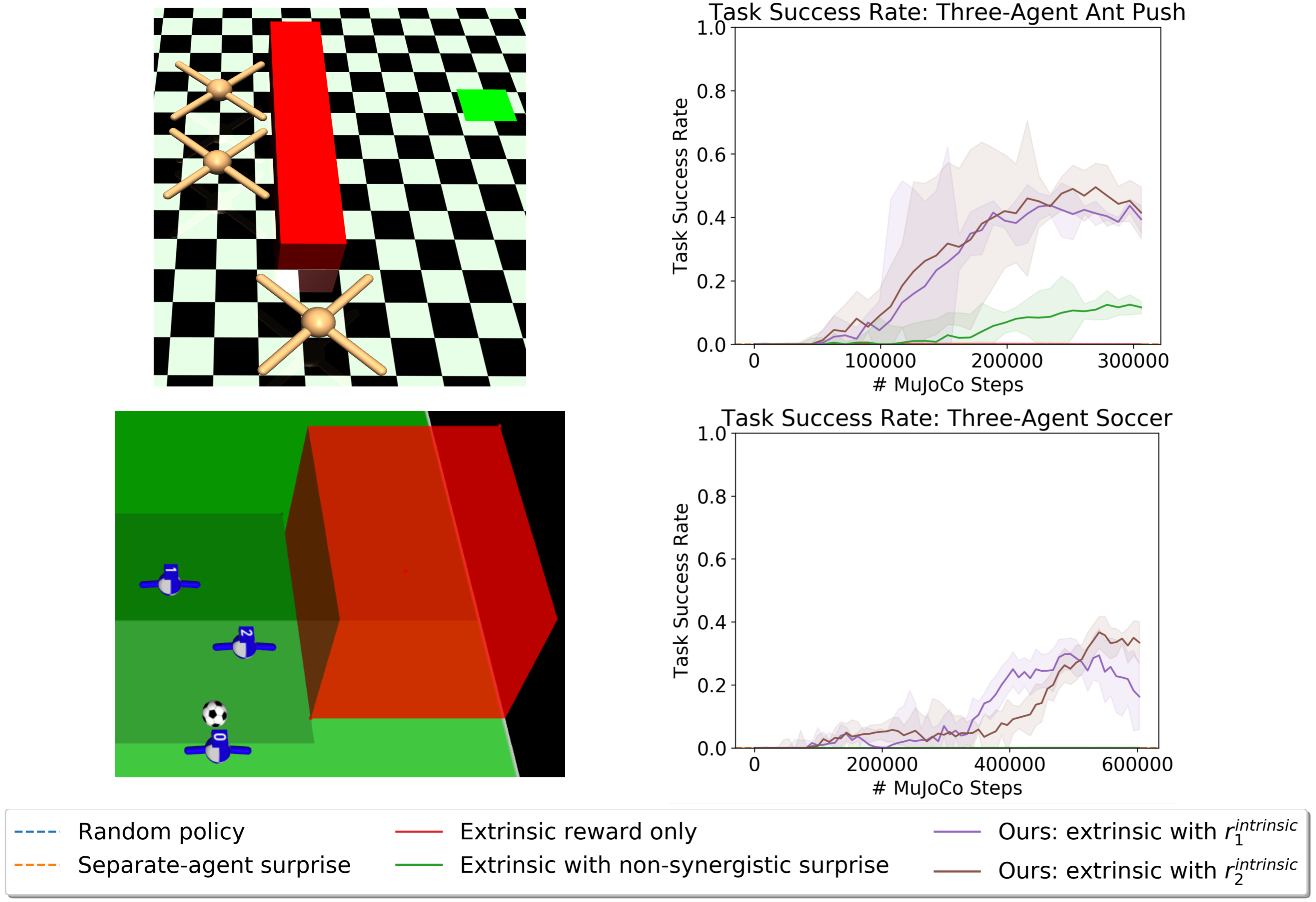}
\end{center}
\caption{\emph{Left}: Screenshots of three-agent versions of ant push
  and soccer environments. \emph{Right}: Learning curves for these
  environments. Each curve depicts an average across 5 random seeds,
  with standard deviations shaded. In these three-agent environments,
  taking random actions almost never leads to success due to the
  exponentially lower likelihood of finding a valid sequence of joint
  actions leading to the goal, and so using only extrinsic reward does
  not perform well. It is apparent that our proposed bias toward
  synergistic behavior is a useful form of intrinsic motivation for
  guiding exploration in these environments as well.}
\label{fig:3agentresults}
\end{figure}

We tested this approach on three-agent versions of the ant
push and soccer environments, and found that it continues to provide a
useful bias. See \figref{fig:3agentresults}. In
three-agent ant push, we give harder goal regions for the ants to push
the blocks to than in two-agent ant push; these regions were chosen by
hand to make all three ants be required to coordinate to solve these
tasks, rather than just two as before. In three-agent soccer, all three
agents must have possessed the ball before the goal is scored.

\section{Conclusion}
\label{sec:conclusion}
In this work, we presented a formulation of intrinsic motivation that
encourages synergistic behavior, and allows efficiently learning
sparse-reward tasks such as bimanual manipulation and multi-agent
locomotion. We observed significant benefits compared to
non-synergistic forms of intrinsic motivation. Our formulation relied
on encouraging actions whose effects would not be achieved by
individual agents acting in isolation. It would be beneficial to
extend this notion further, and explicitly encourage \emph{action
sequences}, not just individual actions, whose effects would not be
achieved by individual agents. Furthermore, while our intrinsic reward
encouraged synergistic behavior in the single policy being learned, it
would be interesting to extend it to learn a diverse set of policies,
and thereby discover a broad set of synergistic skills over the course
of training. Finally, it would be good to extend the domains to
involve more complicated object types, such as asymmetric or
deformable ones; especially for deformable objects, engineering better
state representations is crucial.

\section*{Acknowledgments}
Rohan is supported by an NSF Graduate Research Fellowship. Any
opinions, findings, and conclusions expressed in this material are the
authors' and need not reflect the views of our sponsors.

\bibliography{iclr2020_conference}
\bibliographystyle{iclr2020_conference}

\newpage
\appendix
\section{Pseudocode}
\label{app:pseu}
Here is full pseudocode of our training algorithm described in \secref{sec:rl}:
\begin{algorithm}[h]
  \SetAlgoNoEnd
  \SetAlgoLined
  \DontPrintSemicolon
  \SetKwFunction{algo}{algo}\SetKwFunction{proc}{proc}
  \SetKwProg{myalg}{Algorithm}{}{}
  \SetKwProg{myproc}{Subroutine}{}{}
  \SetKw{Continue}{continue}
  \SetKw{Break}{break}
  \SetKw{Return}{return}
  \myalg{\textsc{Train-Synergistic-Policy}$(\pi_\theta, \M, n,\alpha)$}{
    \nl \textbf{Input:} $\pi_\theta$, an initial policy.\;
    \nl \textbf{Input:} $\M$, an \mdp\ for a synergistic task.\;
    \nl \textbf{Input:} $n$, the number of episodes of data with which to train single-agent models.\;
    \nl \textbf{Input:} $\alpha$, a step size.\;
    \nl \For{$i=1,2,...,n$}{
    \nl Append episode of experience in $\M$ with only agent $A$ acting to data buffer $\data^A$.\;
    \nl Append episode of experience in $\M$ with only agent $B$ acting to data buffer $\data^B$.\;
    }
    \nl Fit forward models $f^A$, $f^B$ to predict next states in $\data^A$, $\data^B$. \tcp*{\footnotesize Pretrained \& fixed.}
    \nl $\data^{\text{joint}} \gets \emptyset$ \tcp*{\footnotesize Data for joint model, only needed if using $r_2^{\text{intrinsic}}$.}
    \nl \While{$\pi_\theta$ has not converged}{
    \nl $\data \gets$ batch of experience tuples $(s_t, a_t, r_t^{\text{extrinsic}}, s_{t+1})$ from running $\pi_\theta$ in $\M$.\;
    \nl \If{using $r_2^{\text{intrinsic}}$}{
        \nl Append $\data$ to $\data^{\text{joint}}$ and fit forward model $f^{\text{joint}}$ to predict next states in $\data^{\text{joint}}$.
    }
    \nl \For{$(s_t, a_t, r_t^{\text{extrinsic}}, s_{t+1}) \in \data$}{
    \nl Replace $r_t^{\text{extrinsic}}$ with $r^{\text{full}}(s_t, a_t, s_{t+1})$. \tcp*{\footnotesize Shape reward, see \secref{sec:rl}.}
    }
    \nl $\nabla_\theta J(\theta) \gets$ \textsc{PolicyGradient}$(\pi_\theta, \data)$\;
    \nl \If{using $r_2^{\text{intrinsic}}$}{
    \nl Update $\nabla_\theta J(\theta)$ with analytical gradients per \eqnref{eq:gradient}.
    }
    \nl $\theta \gets \theta+\alpha \nabla_\theta J(\theta)$ \tcp*{\footnotesize Or Adam~\citep{adam}.}
    }}
\end{algorithm}

\section{Derivation of \eqnref{eq:gradient}}
\label{app:derive}
When using $r_2^{\text{intrinsic}}$, the objective to be optimized can be written as:
\begin{equation*}
    J(\theta) \equiv \Ex_\tau[r^{\text{full}}(\tau)] = \Ex_\tau\left[\sum_{t=0}^T r^{\text{full}}(s_t, a_t, \cdot)\right] = \Ex_\tau\left[\sum_{t=0}^T r_2^{\text{intrinsic}}(s_t, a_t, \cdot)+\lambda\cdot r^{\text{extrinsic}}(s_t)\right].
\end{equation*}

We will write $\nabla_\theta J(\theta)$ in a particular way. Let
$\bar{\tau}_t = \langle s_0, a_0, s_1, a_1, ..., s_t \rangle$ be a
random variable denoting trajectories up to timestep $t$, but
\emph{excluding} $a_t$. We have:
\begin{equation*}
    \begin{split}
        \nabla_\theta J(\theta)&=\nabla_\theta \Ex_\tau[r^{\text{full}}(\tau)]=\sum_{t=0}^T \nabla_\theta \Ex_{\bar{\tau}_t}[\Ex_{a_t \sim \pi_{\theta}(s_t)}[r^{\text{full}}(s_t, a_t, \cdot)]],
    \end{split}
\end{equation*}
where we have used the fact that trajectories up to timestep $t$ have
no dependence on the future $s_{t+1}, a_{t+1}, ..., s_T$, and we have
split up the expectation. Now, observe that the inner expectation,
$\Ex_{a_t \sim \pi_{\theta}(s_t)}[r^{\text{full}}(s_t, a_t, \cdot)]$,
is dependent on $\theta$ since the $a_t$ are sampled from the policy
$\pi_\theta$; intuitively, this expression represents the expected
reward of $s_t$ with respect to the stochasticity in the current
policy. To make this dependence explicit, let us define
$r_\theta^{\text{full}}(s_t) \vcentcolon= \Ex_{a_t \sim
  \pi_{\theta}(s_t)}[r^{\text{full}}(s_t, a_t, \cdot)]$. Then:
\begin{equation*}
\begin{split}
    \nabla_\theta J(\theta)&=\sum_{t=0}^T \nabla_\theta \Ex_{\bar{\tau}_t}[r_\theta^{\text{full}}(s_t)]\\&=\sum_{t=0}^T \int_{\bar{\tau}_t} \nabla_\theta [p_\theta(\bar{\tau}_t)r_\theta^{\text{full}}(s_t)]\ d\bar{\tau}_t\\&=\sum_{t=0}^T \int_{\bar{\tau}_t} p_\theta(\bar{\tau}_t)r_\theta^{\text{full}}(s_t)\nabla_\theta \log p_\theta(\bar{\tau}_t) +p_\theta(\bar{\tau}_t)\nabla_\theta r_\theta^{\text{full}}(s_t)\ d\bar{\tau}_t\\&=\sum_{t=0}^T \Ex_{\bar{\tau}_t}[r_\theta^{\text{full}}(s_t)\nabla_\theta \log p_\theta(\bar{\tau}_t)]+\Ex_{\bar{\tau}_t}[\nabla_\theta r_\theta^{\text{full}}(s_t)],
\end{split}
\end{equation*}
where in the second line, we used both the product rule and the {\sc reinforce} trick~\citep{reinforce}.

Now, let $\tau_t = \langle s_0, a_0, s_1, a_1, ..., s_t, a_t \rangle$ denote trajectories up to timestep $t$, \emph{including} $a_t$ (unlike $\bar{\tau}_t$).
Putting back $\Ex_{a_t \sim \pi_{\theta}(s_t)}[r^{\text{full}}(s_t, a_t, \cdot)]$ in place of $r_\theta^{\text{full}}(s_t)$ gives \eqnref{eq:gradient}:
\begin{equation*}
\begin{split}
 \nabla_\theta J(\theta) &=\sum_{t=0}^T \Ex_{\bar{\tau}_t}[\Ex_{a_t \sim \pi_{\theta}(s_t)}[r^{\text{full}}(s_t, a_t, \cdot)]\nabla_\theta \log p_\theta(\bar{\tau}_t)]+\Ex_{\bar{\tau}_t}[\nabla_\theta \Ex_{a_t \sim \pi_{\theta}(s_t)} [r^{\text{full}}(s_t,a_t,\cdot)]]\\&=\sum_{t=0}^T \Ex_{\tau_t}[r^{\text{full}}(s_t,a_t,\cdot)\nabla_\theta \log p_\theta(\bar{\tau}_t)]+\Ex_{\bar{\tau}_t}[\nabla_\theta \Ex_{a_t \sim \pi_{\theta}(s_t)} [r^{\text{intrinsic}}_2(s_t,a_t,\cdot)]].\qed
 \end{split}
\end{equation*}

In the second line, we have used the facts that $\bar{\tau}_t$ and the
extrinsic sparse reward do not depend on $a_t$. Note that we can
estimate the term
$\Ex_{\bar{\tau}_t}[\nabla_\theta \Ex_{a_t \sim \pi_{\theta}(s_t)}
[r^{\text{intrinsic}}_2(s_t,a_t,\cdot)]]$ empirically using a batch of
trajectory data $\tau^1, ..., \tau^n$, for any timestep $t$.

\section{Additional Environment Details}
\label{app:envdetails}
\subsection{Manipulation Tasks}
We provide additional details about the action space of each manipulation environment.

The following table describes the parameterization of each skill in the library, as well as which environments are allowed to utilize each skill:
\begin{table}[h]
  \centering
  \resizebox{0.8\columnwidth}{!}{
  \tabcolsep=0.08cm{
  \begin{tabular}{c|c|c}
    \toprule[1.5pt]
    \textbf{Skill} & \textbf{Environments} & \textbf{Continuous Parameters}\\
    \midrule[2pt]
    top grasp & bar, ball, bottle & end effector position, end effector z-orientation\\
    \midrule
    side grasp & bottle, corkscrew & end effector position, approach angle\\
    \midrule
    go-to pose & ball, corkscrew & end effector position, end effector orientation\\
    \midrule
    lift & bar, ball & vertical distance to lift end effector\\
    \midrule
    twist & bottle & none (wrist joint rotates at current end effector pose)\\
    \midrule
    rotate & corkscrew & rotation axis, rotation radius\\
    \midrule
    no-op & all & none\\
    \bottomrule[1.5pt]
  \end{tabular}}}
\end{table}

The following table describes the search space of each continuous parameter. Since the object pose is known in simulation, we are able to leverage it in designing these search spaces:
\begin{table}[h]
  \centering
  \resizebox{\columnwidth}{!}{
  \tabcolsep=0.08cm{
  \begin{tabular}{c|c|c|c}
    \toprule[1.5pt]
    \textbf{Continuous Parameter} & \textbf{Environments} & \textbf{Relevant Skills} & \textbf{Search Space}\\
    \midrule[2pt]
    end effector position (unitless) & bar & top grasp & [-1, 1] interpolated position along bar \\
    \midrule
    end effector position (meters) & ball, bottle, corkscrew & grasps, go-to pose & [-0.1, 0.1] x/y/z offset from object center\\
    \midrule
    end effector z-orientation & bar, ball, bottle & top grasp & $[0, 2\pi]$ \\
    \midrule
    approach angle & bottle, corkscrew & side grasp & $[-\frac{\pi}{2}, \frac{\pi}{2}]$ \\
    \midrule
    end effector orientation & ball, corkscrew & go-to pose & $[0, 2\pi]$ r/p/y Euler angles converted to quat\\
    \midrule
    distance to lift (meters) & bar, bottle & lift & $[0, 0.5]$ \\
    \midrule
    rotation axis & corkscrew & rotate & [-0.1, 0.1] x/y offset from object center; vertical \\
    \midrule
    rotation radius (meters) & corkscrew & rotate & $[0, 0.2]$ \\
    \bottomrule[1.5pt]
  \end{tabular}}}
\end{table}

Note that our inverse kinematics feasibility checks allow the system to learn to
rule out end effector poses which are impossible to reach, since these
cause no change in the state other than consuming a timestep, and
generate 0 reward.

\subsection{Locomotion Tasks}
We provide additional details about the action space of the locomotion
environments. For both the ant push and soccer tasks, we
follow~\citet{diversification} and pre-train four skills: moving up,
down, left, and right on the plane. Each skill has one continuous
parameter specifying an amount to move. So, at each timestep, the
policy must select both which direction to move and how much to move
in that direction. All training hyperparameters are unchanged from the
manipulation tasks.

\subsection{Policy Architecture}
\begin{figure}[h]
\begin{center}
\includegraphics[width=0.6\linewidth]{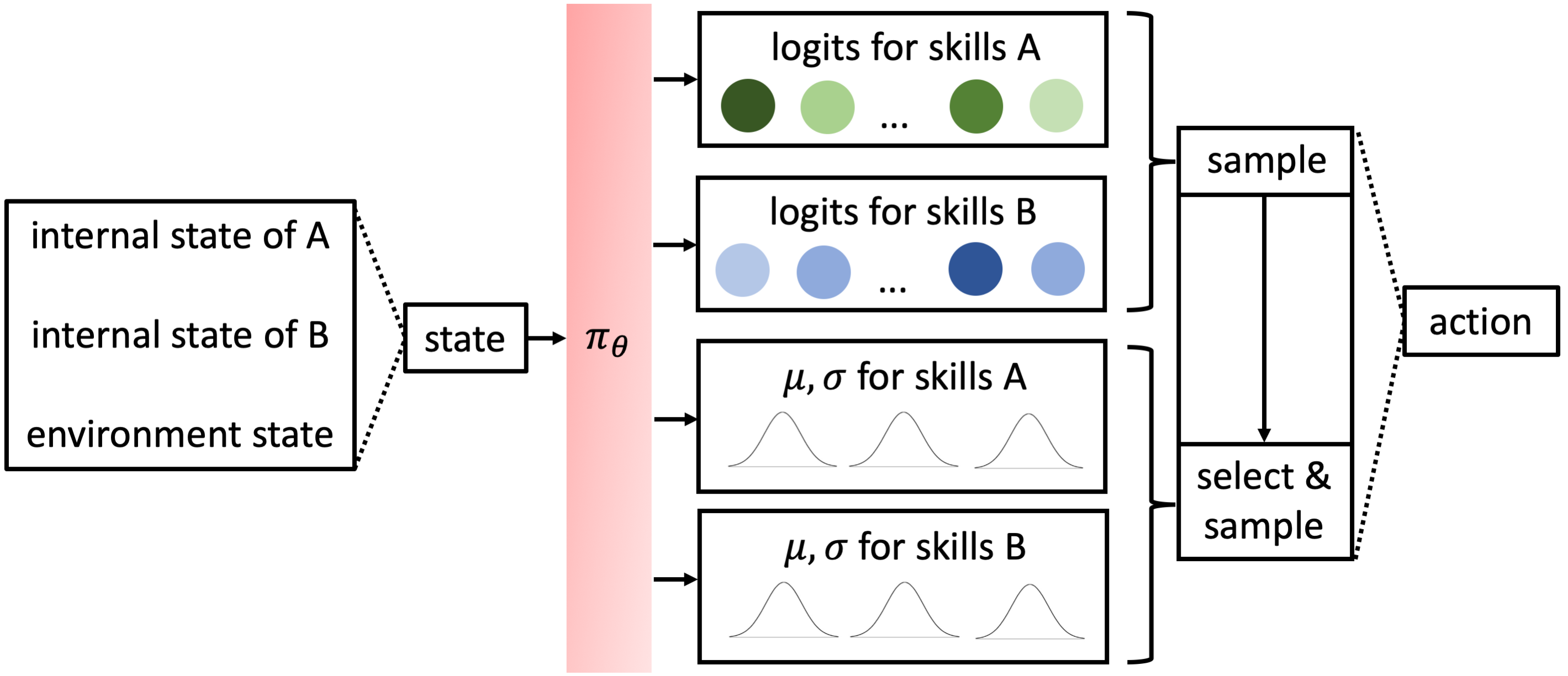}
\end{center}
\caption{The policy $\pi_\theta$ maps a state to 1) a categorical
  distribution over skills for $A$, 2) a categorical distribution over
  skills for $B$, 3) means and variances of independent Gaussian
  distributions for every continuous parameter of skills for $A$, and
  4) means and variances of independent Gaussian distributions for
  every continuous parameter of skills for $B$. To sample from the
  policy, we first sample skills for $A$ and $B$, then sample all
  necessary continuous parameters for the chosen skills from the
  Gaussian distributions. Altogether, the two skills and two sets of
  parameters form an action, which can be fed into the forward models
  for prediction.}
\label{fig:model}
\end{figure}

\figref{fig:model} shows a diagram of our policy architecture.

\section{Impact of Coefficient $\lambda$}
\label{app:lambda}
\begin{figure}[h]
\begin{center}
\includegraphics[width=0.5\linewidth]{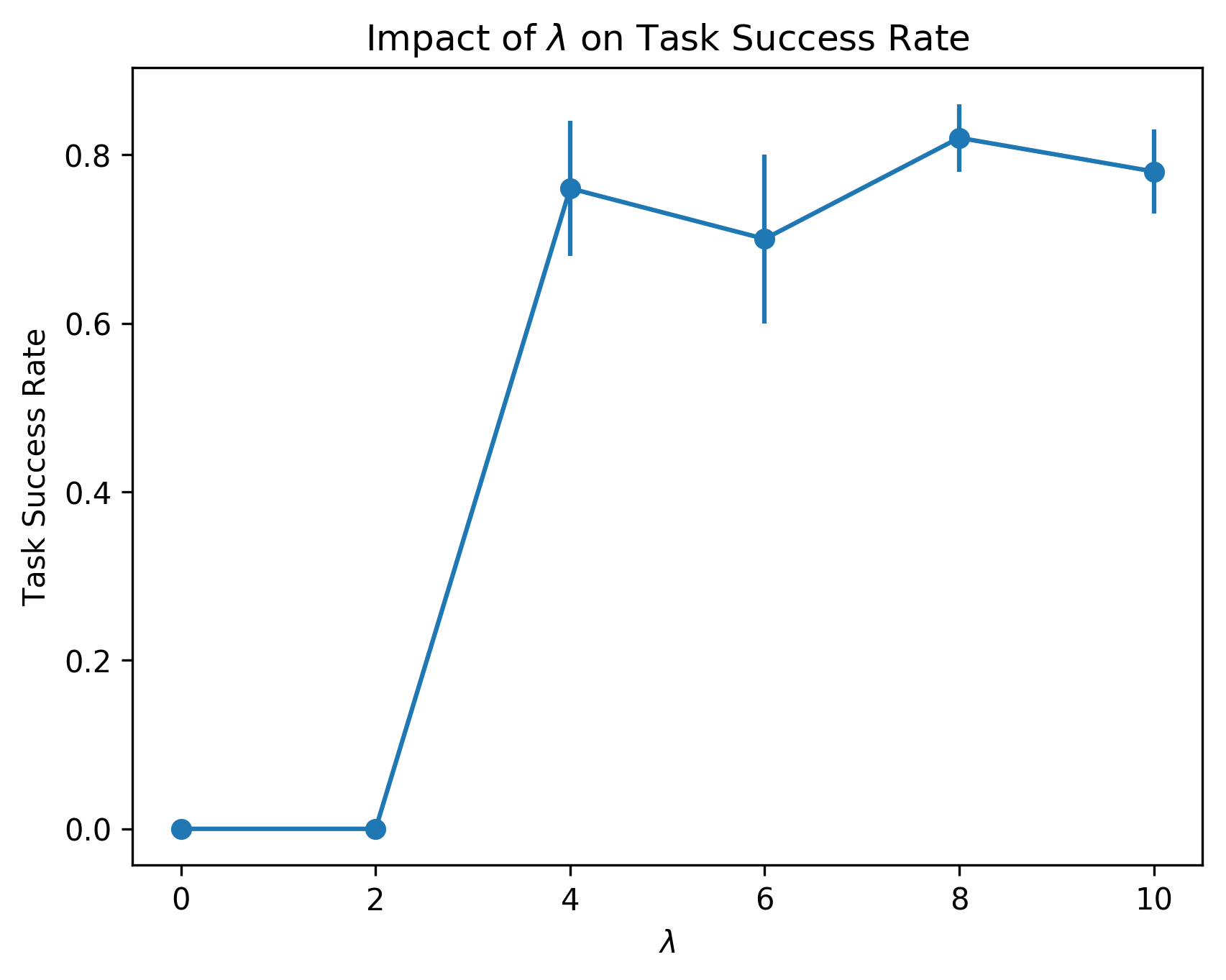}
\end{center}
\caption{Task success rate of learned policy at convergence for the
  bar pickup task, using our best-performing reward function
  $r_2^{\text{intrinsic}}$. Each result is averaged across 5 random
  seeds, with standard deviations shown. We can infer that once
  $\lambda$ reaches a high enough value for the extrinsic rewards to
  outscale the intrinsic rewards when encountered, the agents will be
  driven toward behavior that yields extrinsic rewards. These
  extrinsic, sparse rewards are only provided when the task is
  successfully completed.}
\label{fig:lambda}
\end{figure}

We conducted an experiment to study the impact of the trade-off
coefficient $\lambda$ on the performance of the learned policy. When
$\lambda=0$, no extrinsic reward is used, so the agents learn to act
synergistically, but in ways that do not solve the ``task,'' which is
sensible since the task is unknown to them. Our experiments reported
in the main text used $\lambda=10$. See \figref{fig:lambda} for the
results of this experiment.

\end{document}